%% file: submission.tex
\pdfoutput=1

\documentclass[11pt]{article}

\usepackage[preprint]{acl}

\usepackage{times}
\usepackage{latexsym}

\usepackage[T1]{fontenc}

\usepackage[utf8]{inputenc}

\usepackage{microtype}

\usepackage{inconsolata}

\usepackage{graphicx}

\usepackage{listings}
\usepackage{xspace}
\usepackage{booktabs}
\usepackage{array}
\usepackage{enumitem}
\usepackage{xspace}

\usepackage{xcolor}
\usepackage{pifont}

\newcommand{\yes}{\textcolor{green!60!black}{\ding{51}}}
\newcommand{\no}{\textcolor{red!70!black}{\ding{55}}}

\input{commands}

\newcommand{\framework}{\textsc{CQA-Eval}\xspace}

\title{\framework: Designing Reliable Evaluations of Multi-paragraph Clinical QA under Resource Constraints}

\author{
 \textbf{Federica Bologna\textsuperscript{1}},
 \textbf{Tiffany Pan\textsuperscript{1}},
 \textbf{Matthew Wilkens\textsuperscript{1}},
 \textbf{Yue Guo\textsuperscript{2}},
 \textbf{Lucy Lu Wang\textsuperscript{3}}
\\
 \textsuperscript{1}Cornell University,
 \textsuperscript{2}University of Illinois, Urbana-Champaign,
 \textsuperscript{3}University of Washington,
\\
 \small{
   \textbf{Correspondence:} \href{mailto:fb265@cornell.edu}{fb265@cornell.edu}
 }
}

\begin{document}
\maketitle
\begin{abstract}
Evaluating multi-paragraph clinical question answering (QA) systems is resource-intensive and challenging: accurate judgments require medical expertise and achieving consistent human judgments over multi-paragraph text is difficult. We introduce \framework, an evaluation framework and set of evaluation recommendations for limited-resource and high-expertise settings. Based on physician annotations of 300 real patient questions answered by physicians and LLMs, we compare coarse answer-level versus fine-grained sentence-level evaluation over the dimensions of correctness, relevance, and risk disclosure.
We find that inter-annotator agreement (IAA) varies by dimension: fine-grained annotation improves agreement on correctness, coarse improves agreement on relevance, and judgments on communicates-risks remain inconsistent. 
Additionally, annotating only a small subset of sentences can provide reliability comparable to coarse annotations, reducing cost and effort.\footnote{Data and code available at \href{https://anonymous.4open.science/r/6C5C/}{https://anonymous.4open.sci-ence/r/6C5C/} under a CC-BY license.}

\end{abstract}

\section{Introduction}

\begin{table*}[t]
\centering
\small
\setlength{\tabcolsep}{4pt}
\begin{tabular}{l c l l c c c c c c}
\toprule
\textbf{\shortstack[l]{Dataset\\(author, year)}} & \textbf{\shortstack[c]{\# QA\\pairs}} & \textbf{\shortstack[l]{Question\\source}} & \textbf{\shortstack[l]{Reference\\source}} & \textbf{\shortstack[c]{Number\\raters}} & \textbf{\shortstack[c]{Number\\criteria}} & \textbf{IAA} & \textbf{\shortstack[c]{Expert\\ratings}} & \textbf{Codebook} & \textbf{\shortstack[c]{LLM-as-\\judge}} \\
\midrule
Ben Abacha et al.~\citeyear{ben_abacha_question-entailment_2019} & 2{,}479 & Web & Extracted & 3 & 1 & \yes & \yes & \no & \no \\
Ayers et al.~\citeyear{ayers_comparing_2023} & 100 & Reddit & Extracted & 3 & 3 & \no & \yes & \no & \no \\
Singhal et al.~\citeyear{singhal_large_2023} & 140 & Mix & Experts & 1 & 12 & \no & \no & \no & \no \\
Singhal et al.~\citeyear{singhal_towards_2023} & 1{,}446 & Mix & Experts & 15 & 12 & \yes & \no & \no & \no \\
Our study & 300 & Health app & Experts & 6 & 3 & \yes & \yes & \yes & \yes \\
\bottomrule
\end{tabular}
\caption{Comparison of clinical QA evaluation datasets and annotation practices.}
\label{tab:dataset_comparison}
\end{table*}
Rising healthcare costs and limited provider availability make it difficult for patients to receive timely answers to their clinical questions. Generative models integrated into electronic health record (EHR) systems could help, but evaluating their responses requires medical expertise \cite{ayers_comparing_2023, singhal_large_2023}. Expert annotators are costly, scarce, and often disagree on what constitutes a good answer, leading to low inter-annotator agreement (IAA) \cite{hosseini_benchmark_2024, leonardelli_agreeing_2021, khasentino_personal_2025}. Simplifying the annotation task is thus fundamental to reducing annotation fatigue and improving annotation reliability.

Although fine-grained sentence-level annotations improve IAA on faithfulness in summarization~\cite{krishna_longeval_2023}, it remains unclear whether these benefits extend to other dimensions, particularly in high-stakes domains where disagreement is common~ \cite{leonardelli_agreeing_2021, mostafazadeh_davani_dealing_2022, hosseini_benchmark_2024}. 
In this work, we validate these guidelines in the high-expertise, high-consequence setting of consumer health QA and examine their generalizability to annotation facets beyond faithfulness.

We introduce \framework, a framework for evaluating multi-paragraph clinical QA across the dimensions of \emph{correctness}, \emph{relevance}, and \emph{communicates-risks}. 
We conduct a randomized human annotation study on pairs of patient questions and clinical answers, tasking annotators with evaluating answers under two annotation designs: a \textit{coarse} design in which raters assess the entire answer, and a \textit{fine-grained} design in which they evaluate individual sentences. This allows us to assess whether specific annotation designs can reduce annotator fatigue while maintaining or improving annotation quality. We also assess the viability of partial annotations, the impact of annotation design on system ratings,
and whether our annotation designs can be adapted for LLM-as-judge settings.

Based on our annotation study findings, we derive a set of recommendations for achieving reproducible annotations under resource and expertise constraints. 
We find, for example, that fine-grained annotations improve IAA primarily for factual dimensions such as correctness, while coarse annotations are sufficient or sometimes better for dimensions like \emph{relevance} and \emph{communicates-risks}. At the same time, partial fine-grained annotation---evaluating only a few sentences per answer---can achieve ratings and variance comparable to full fine-grained ratings while reducing annotation cost. These results suggest that annotation design should be tailored to the evaluation dimension: fine-grained schemes for factuality-based dimensions, coarse ones for context-based ones.

Our LLM-as-judge experiments show that fine-grained instructions are less effective at improving LLM-expert agreement than expert-expert agreement. Nonetheless, our annotation instructions yield LLM-expert agreement comparable to expert-expert agreement for correctness and relevance. In addition, we find that fine-grained evaluation may help counteract length-related biases of coarse assessments.

In practice, fine-grained ratings for correctness and coarse ratings for relevance provide reliable, efficient metrics for assessing LLM-generated clinical answers.
In sum, our contributions include:
\begin{enumerate}[itemsep=-2pt, topsep=0pt, leftmargin=12pt]
    \item A \textbf{dataset} of 300 question-answer pairs with annotations for correctness, relevance, and risk disclosure collected from six independent medical experts; 
    \item The \framework \textbf{annotation framework} for both coarse and fine-grained evaluation of answers to clinical questions;
    \item A \textbf{comparison study} of coarse and fine-grained annotations through a randomized human annotation study of ground-truth answers provided by clinicians and those generated by two LLMs widely used for this task; 
    \item A set of \textbf{recommendations} that helps developers of clinical LLMs find the best annotation design for their goals;
    \item An \textbf{analysis} of how our annotation framework generalizes to LLM-as-judge settings.
\end{enumerate}

\input{02-related-work}

\section{Annotation Study Design}
To evaluate reliable designs for multi-paragraph clinical QA, we augment a dataset of real patient questions with LLM-generated answers (\S\ref{dataset}), define evaluation dimensions and granularities (\S\ref{task_formulation}), and recruit experienced physicians to provide annotations (\S\ref{experimental_design}).

\subsection{Dataset} \label{dataset}
We conduct experiments on the K-QA dataset \cite{manes_k-qa_2024}, which consists of questions written by real patients who use K Health, an AI-driven clinical platform with over 8 million users. These consumer health questions span general primary care topics, from medication safety to disease prevention, covering hundreds of medical conditions from neurology to reproductive health. 
Reference answers in the dataset are curated by a team of medical doctors. 

We randomly sample 100 questions and their corresponding physician answers from K-QA for our annotation study. 
We augment these QA pairs by generating answers to the patient questions using GPT-4 \cite{openai_gpt-4_2024} and Llama-3.1-Instruct-405B \cite{grattafiori_llama_2024}. We focus on these two models because they are widely deployed and have shown strong performance on medical tasks in previous studies \cite{kanithi2024medic}. The models are prompted using five-shot in-context learning (ICL) \citep{Brown2020LanguageMA} and chain-of-thought reasoning \citep{Wei2022ChainOT}. The full prompt is available on \href{https://anonymous.4open.science/r/6C5C/}{Github}. 
In each case, we instruct the model to limit its answers to 270 words, the maximum length of physician answers in the K-QA dataset, and set the max number of output tokens to 300. Four answers generated by Llama-3.1-Instruct-405B still exceeded this limit and are returned with truncated sentences; we remove these from the analysis. Table~\ref{tab:token_stats} shows answer length distributions; both models produce longer answers on average than the original physician answers. 

We provide example model and physician answers in Appendix~\ref{sec:kqa-examples}. 
Each of the 100 sampled patient questions is paired with three answers (physician, GPT-4, Llama), yielding 300 total QA pairs. 

\begin{table}[t!]
\centering
\begin{tabular}{lc}
\toprule
\textbf{Answer type} & \textbf{Mean tokens ± SD} \\
\midrule
GPT-4 & 124 ± 50 \\
Llama-3.1-Instruct-405B & 170 ± 52 \\
Physician (ground-truth) & 106 ± 54 \\
\bottomrule
\end{tabular}
\caption{Average number of tokens and standard deviation for each answer type in our dataset.}
\label{tab:token_stats}
\end{table}

\subsection{Annotation Task Formulation} \label{task_formulation}
There is no consensus on which dimensions of clinical answers should be assessed. After carefully reviewing codebooks and rubrics of previous evaluation studies of clinical datasets \cite{singhal_large_2023,singhal_towards_2023,hosseini_benchmark_2024,ayers_comparing_2023,ben_abacha_question-entailment_2019, ben_abacha_investigation_2023, kanithi_medic_2024}, we identify three dimensions:

\begin{itemize}[noitemsep, topsep=0pt, leftmargin=10pt]
    \item \textbf{Correctness}: the answer aligns with current medical knowledge
    \item \textbf{Relevance}: the answer addresses the specific medical question asked
    \item \textbf{Communicates-risks}: the answer communicates contraindications or risks 
\end{itemize}

We decide to evaluate answers on risk disclosure rather than possibility of harm because it is complementary to correctness. Both of these dimensions (omitting mentions of potential risks and providing incorrect medical information) have the potential to harm the reader. More detailed and extensive definitions of these dimensions can be found in Appendix \ref{sec:dimensions}.

In our study, annotators evaluate multi-paragraph answers to patient questions along these three dimensions under two annotation settings:
\begin{itemize}[itemsep=-2pt, topsep=0pt, leftmargin=10pt]
    \item \textbf{Coarse-grained}: annotators are shown a question and answer and asked to consider the \emph{answer as a whole}; they rate each dimension of the answer on a five-point Likert scale---\textit{Disagree, Partially Disagree, Neutral, Partially Agree, Agree}.
    \item \textbf{Fine-grained}: annotators are shown a question and answer with one sentence highlighted, and asked to consider the \emph{sentence} in the context of the whole answer; they assess each dimension of the sentence on a five-point Likert scale. 
\end{itemize}

\noindent Prior work on fine-grained annotations explored a variety of units, such as entities, numbers, clauses, or sentences \cite{krishna_longeval_2023}. We chose to conduct fine-grained evaluation at the sentence-level because answers in the K-QA dataset are generally concise and annotating entities would be prohibitively expensive. As 75\% of answers consist of fewer than eight sentences, we randomly sample six sentences 
from each answer to be evaluated, balancing annotation cost with coverage. We hypothesize that dimensions where evidence is localized (e.g., correctness of a factual claim) would benefit most from fine-grained annotations.

We develop two codebooks, one for each setting, providing annotators with detailed definitions and criteria, as well as instructions and real-world examples. Information in our codebooks are based on articles drawn from reputable sources such as Mayo Clinic, Cleveland Clinic, and Harvard Medical School. The full codebooks can be found on \href{https://anonymous.4open.science/r/6C5C/}{Github}. Annotators have access to the codebook at all times while performing the annotation task.

\subsection{Experimental Design} \label{experimental_design}

\paragraph{Annotators} We hire six physicians from the freelance recruiting platform Upwork following previous work \cite{yim_morqa_2025, jin_better_2023}. We select candidates based on their credentials and their recent jobs on the platform. Their clinical expertise was further determined through two pilot studies where they were asked to provide motivations in case of large disagreement with other annotators.

All the selected physicians have first-hand experience with patient care. After providing consent, they reported specialties in General Medicine, Internal Medicine, Pathology, Radiology, or Surgery and had between 3 and 15 years of patient-care experience.
Annotators are compensated at a rate of 25 USD per hour for their work.

\paragraph{Annotation Design} We divide the six annotators into two groups, where \textit{Group A} is assigned all answers (from the physician and both LLMs) for 50 randomly sampled patient questions and \textit{Group B} is assigned all answers for the remaining 50 patient questions. In other words, each annotator evaluates 150 QA pairs, and each QA pair is evaluated by 3 physicians. 
For each annotator, half of these assigned QA pairs are annotated
following a coarse annotation design, and half following a fine-grained annotation design. To implement this, each group’s 150 pairs are split evenly into 17 coarse batches and 17 fine-grained batches, with 9 pairs per batch except for batch 17, which contains 6 pairs. The order of questions within each batch is randomized to control for confounding due to fatigue. For each annotator, we also randomize in which batch each of the 50 questions appears, to control for temporal effects in annotation decisions. However, to reduce context switching, all three possible answers for each question are grouped and always assigned to the same batch. 

To minimize bias introduced by the sequence in which the two annotation designs are performed, we alternate coarse and fine-grained annotation tasks. Annotators first complete ten coarse batches, followed by eight fine-grained batches with entirely new pairs. They then evaluate ten fine-grained batches on pairs they had previously annotated under the coarse scheme, and conclude with the remaining eight coarse batches, which include pairs they had already evaluated in the fine-grained setting. Lastly, annotators repeat annotations for 6 QA pairs in both coarse and fine-grained settings that they had evaluated at the beginning of the study, allowing us to measure intra-rater reliability (IRR) for each annotator.

\paragraph{Additional Measures}
To measure \emph{annotator effort}, we record the time elapsed between when annotators are shown a new QA pair and when they proceed to the next one. For each QA pair, we also measure \emph{annotator confidence}. We ask annotators to respond to the question “How confident do you feel about your evaluation of the answers?” on a five-point Likert scale ranging from \textit{Not confident at all} to \textit{Very confident}.  

At the end of each annotation batch (9 QA pairs), we collect feedback on the clarity of the codebook instructions. Annotators rate “How easy to follow are the annotation instructions?” on a five-point Likert scale ranging from \textit{Very difficult} to \textit{Very easy}.

\paragraph{Post-annotation Survey}
At the end of the annotation study, annotators complete a two-part feedback survey structured after the NASA Task Load Index (TLX) \cite{hart_development_1988}. The first part asks about their experience in the coarse annotation setting, while the second repeats the same questions for the fine-grained setting. Questions are reproduced in Appendix~\ref{sec:nasatlx}; annotators answer each question on a 7-point scale, allowing us to measure and compare annotators’ perceived effort across both task designs. We also collect qualitative feedback on preferences between annotation settings, asking the open-ended question ``Which annotation design do you prefer?'' 

\subsection{LLM-as-judge}
Limited evidence exists regarding the ability of LLMs to provide human-aligned evaluations of model-generated clinical answers. \citet{jin_better_2023} show that expert annotators agree 75\% of the time with GPT-4o regarding the correctness of clinical answers. However, \citet{szymanski_limitations_2025} find a much lower agreement rate in the context of dietetics and mental health. We fill this gap by assessing our evaluation dimensions with LLM-judges, and reporting the IAA between the LLM-judge and our expert annotators. We investigate whether LLM-as-judge agreement with expert raters is impacted by coarse- versus fine-grained prompting strategies.

Following prior work, we select GPT-4o as our judge model \cite{chen_benchmarking_2025, hosseini_benchmark_2024}.
We prompt the model with our two codebooks, each minimally adapted from the human versions. Full prompts can be found on \href{https://anonymous.4open.science/r/6C5C/}{Github}. 

\subsection{Measures of Reliability}
To ensure comparability across settings, we collapse coarse annotations from a 5-point to a 3-point Likert scale. Fine distinctions (e.g., Agree vs. Partially Agree) introduced variability without added value, consistent with prior work showing that collapsing categories improves reliability in subjective tasks \cite{artstein2008inter, davani2022dealing}. We map ratings as follows: \texttt{+1} for Agree/Partially Agree, \texttt{0} for Neutral, and \texttt{-1} for Disagree/Partially Disagree. 
For fine-grained annotations, we consider two schemes. In Fine 3pt, we collapse 5-point ratings to 3-point at the sentence level and report IAA. In Fine Binary, sentence-level ratings are aggregated into a passage-level judgment, reflecting dimension-specific risks: any incorrect sentence invalidates correctness (\texttt{0}), while at least one sentence that is relevant or communicates risks ensures adequacy (\texttt{1}). This binary scheme enables fairer comparison with coarse annotations and yields more interpretable system-level outcomes.

We use IAA as our primary reliability measure, reporting Randolph’s $\kappa$, which better captures agreement among experts than Fleiss' $\kappa$ or Krippendorff's $\alpha$~\citet{singhal_towards_2023}. For repeated annotations, we measure IRR as percent self-agreement. We also report annotators’ self-rated confidence and ease (average Likert scores) and effort (average seconds per QA pair, aggregated across sentences in fine-grained settings).



\section{Results \& Recommendations}

\begin{figure}[t]
  \includegraphics[width=\columnwidth]{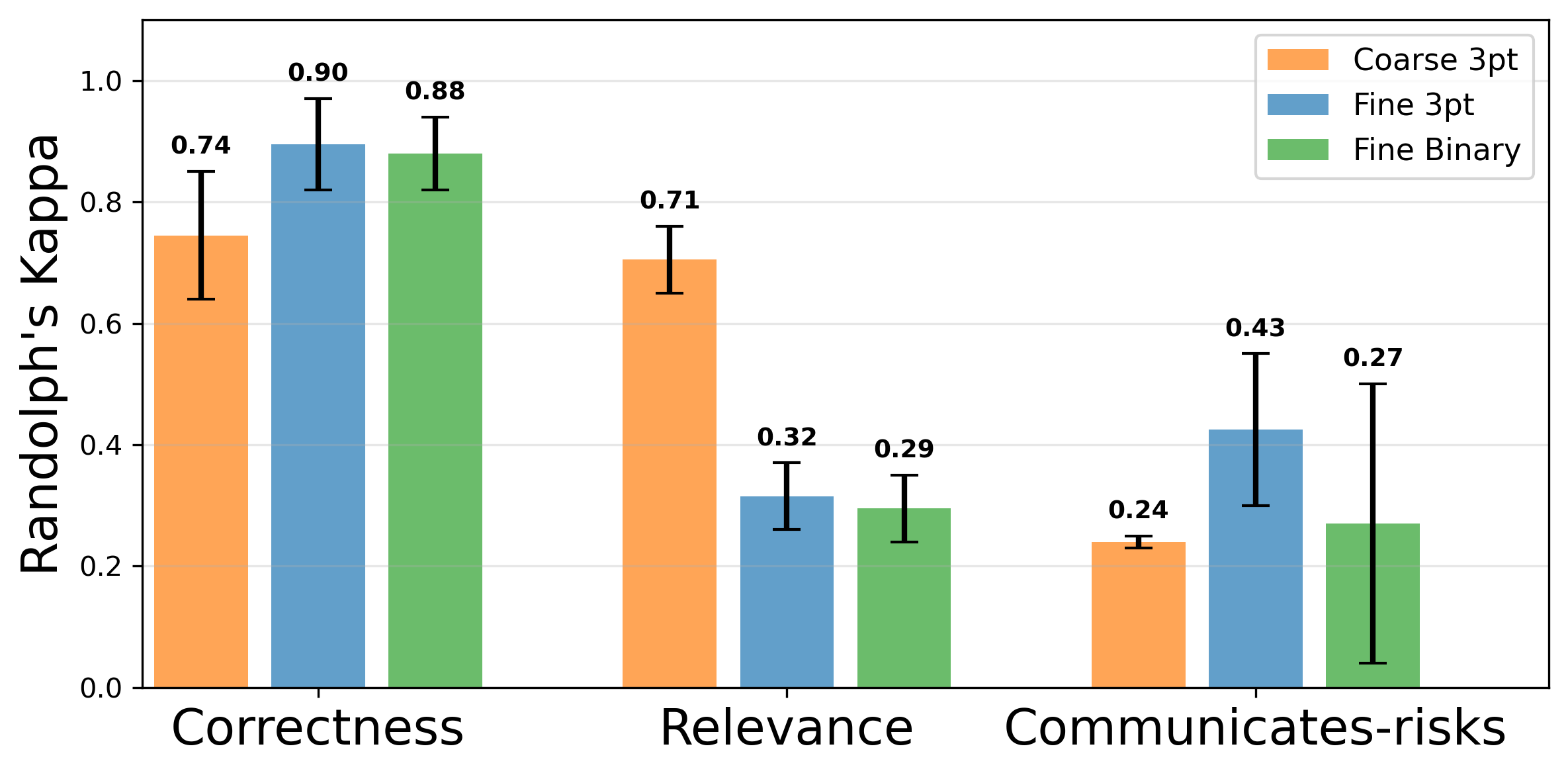}
  \caption{Average inter-annotator agreement (IAA) (Randolph’s $\kappa$) for expert annotators across all annotation groups for the three dimensions: correctness, relevance, and communicates-risks. 
  }
  \label{fig:iaa}
\end{figure}

\paragraph{Fine-grained annotations achieve higher IAA for correctness but not for other dimensions.}
For correctness, agreement is substantially higher in the fine-grained setting (0.90, 0.88) compared to the coarse 3-point setting (0.74) (Figure \ref{fig:iaa}). In contrast, annotators reached higher agreement in the coarse setting (0.71) than in the fine-grained settings (0.32, 0.29) in the case of relevance. Communicates-risks shows low agreement in both settings, with an increase in the fine-grained settings (0.43, 0.27) compared to the coarse binary setting (0.24). Because IAA improvement in fine-grained settings varies based on the aggregation strategy, we cannot reliably conclude that fine-grained annotations improve IAA for communicates-risks.

The first part of our results confirm findings from \citet{krishna_longeval_2023} that IAA among annotators increases with fine-grained annotations. However, we find this only holds true for specific dimensions, notably correctness. We hypothesize that dimensions that rely on factuality and the alignment of a single piece of text against a reference---such as faithfulness (as assessed in \citet{krishna_longeval_2023}) and correctness---are more suited to a fine-grained annotation design. Dimensions that are constructed across multiple pieces of texts (such as relevance), or that rely on identifying missing information (such as communicates-risks) may not be efficiently captured by fine-grained annotations.

This hypothesis is supported by annotator responses to our feedback survey. They reported that, while fine-grained (referred to as \textit{sentence-level} in the survey) annotations allowed them to provide more accurate ratings, those annotations are more difficult in cases where context is important. A sample of quotes from annotator responses is available in Appendix \ref{sec:feedback-survey}. Furthermore, responses on the NASA-TLX (Table~\ref{tab:additional-measures}) show that annotators, on average, found coarse annotations to be easier and less mentally and physically demanding. However, they felt more successful in completing annotations and less discouraged when following fine-grained instructions.

In addition to comparing IAA across the two annotation settings, we also measure IRR in both coarse and fine-grained settings on 6 QA pairs
experts evaluated at the beginning and end of the study. We report these results in Appendix~\ref{sec:IRR}. Given the small sample size, these results did not achieve statistical significance and do not offer strong evidence for recommendations. Future work should investigate whether a fine-grained design improves IRR across dimensions.

\textbf{\textit{Recommendation}}: Researchers should choose an annotation design based on the dimensions being assessed. Developing a detailed codebook, and collecting annotator feedback through a pilot, survey, or interview before starting the evaluation study can help researchers decide whether coarse or fine-grained annotations are better suited for the dimension of interest.

\paragraph{Partial fine-grained annotations achieve comparable results at half the cost.}

\begin{figure*}[t]
  \includegraphics[width=0.48\linewidth]{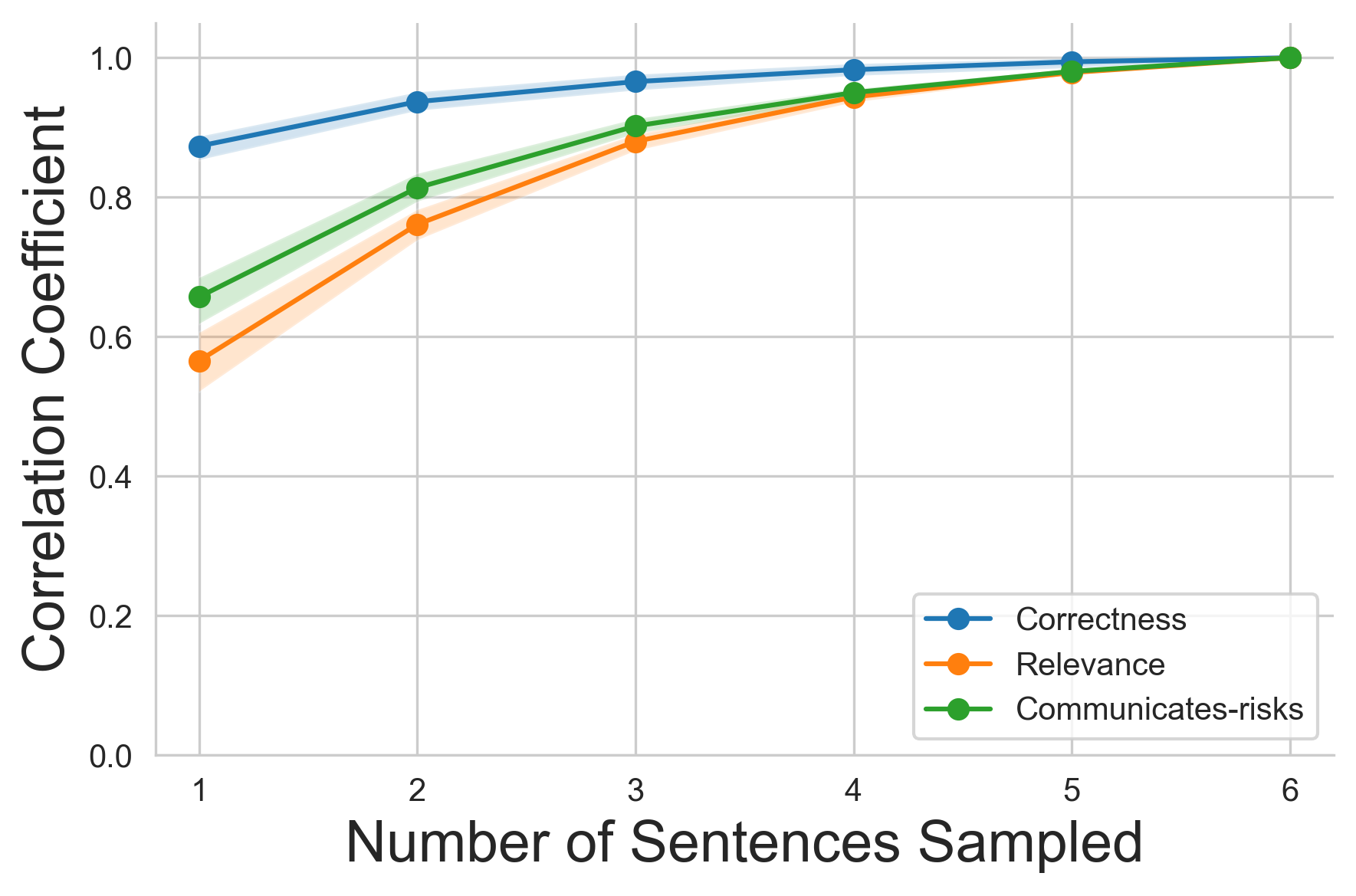} \hfill
  \includegraphics[width=0.48\linewidth]{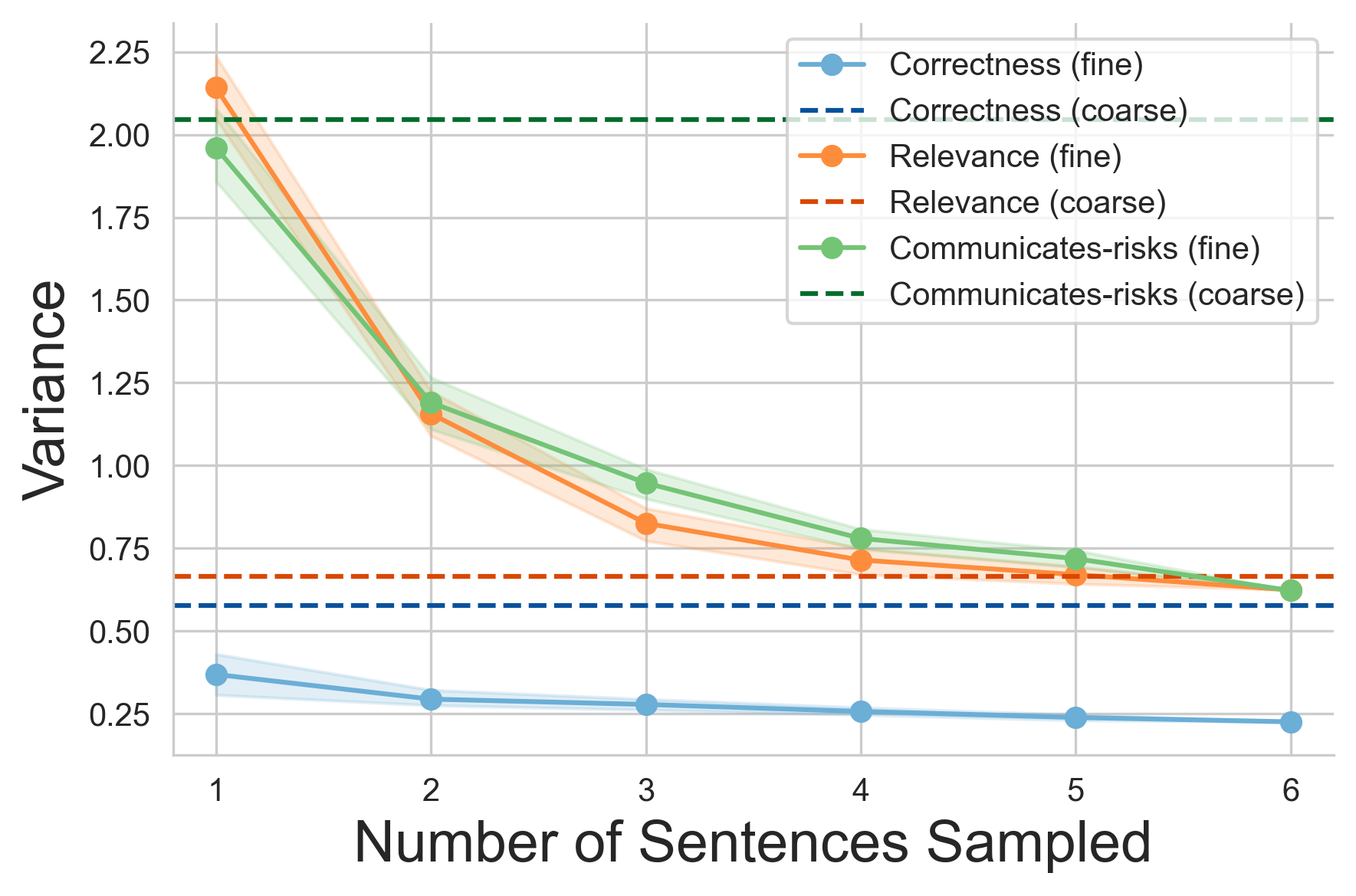}
  \caption {Correlation between partial fine-grained annotations and full fine-grained annotations and variance when partially annotating an answer (left). Inter-annotator variance when partially annotating an answer compared against coarse annotations. Confidence intervals shown are 95\% and computed across 100 random subsets.}
  \label{fig:sampled}
\end{figure*}

Calculating Spearman's rank correlation coefficient between partial and full fine-grained annotations, we find that ratings based on assessing only three sentences per answer already correlate strongly (correlation coefficient above 0.8) with six-sentence annotations (Figure~\ref{fig:sampled} left). Furthermore, variance in fine-grained ratings decreases rapidly with additional sentences: with only three sentences, variance is lower than coarse annotations for correctness and communicates-risks, and only slightly higher for relevance (Figure~\ref{fig:sampled} right).

Although coarse annotations are significantly faster---taking an average of 239.3 seconds per answer compared to 459.8 seconds for full fine-grained annotations (Table~\ref{tab:additional-measures}; p<0.001, effect size of $-0.454$)---sampling three sentences reduces the annotation time and cost to a level comparable with coarse annotations while maintaining the same IAA.
In terms of confidence, annotators report similar ratings for coarse (mean = 3.37) and fine-grained (mean = 3.40) annotations, with no significant difference (p = 0.5423). 

\begin{table}[t!]
\centering
\begin{tabular}{lccc}
\toprule
\textbf{Measure} & \textbf{Coarse} & \textbf{Fine} & \textbf{$p$-value} \\
\midrule
Confidence & 3.37 & 3.40 & $=$ .54 \\
Ease & 3.05 & 2.56 & $<$ .001 \\
Time spent (sec) & 239.3 & 459.8 & $<$ .001 \\
\bottomrule
\end{tabular}
\newline
\vspace{2mm}
\newline
\centering
\begin{tabular}{lcc}
\toprule
\textbf{Question} & \textbf{Coarse} & \textbf{Fine} \\
\midrule
Mental demand & 4.00 & 4.50 \\
Physical demand & 3.67 & 4.33 \\
Temporal demand & 3.83 & 3.50 \\
Performance & 6.00 & 6.33 \\
Effort & 4.00 & 4.83 \\
Frustration & 2.50 & 1.50 \\
\bottomrule
\end{tabular}
\caption{Above: Comparison of annotator self-reported confidence, ease, and time taken to complete tasks between coarse and fine-grained annotation settings. Below: Mean NASA-TLX ratings for coarse and fine-grained tasks. Ratings are reported on a 7-point scale where higher indicates more of the thing in question.}
\label{tab:additional-measures}
\end{table}


\textbf{\textit{Recommendation}}: Sampling a portion of the answer 
can be a cost-effective strategy for fine-grained evaluation. We find that this preserves reliability and variance relative to larger samples, while keeping annotation costs and annotator effort comparable to coarse annotations. 

\paragraph{Fine-grained annotations may mitigate biases related to response length when computing system-level ratings and rankings}

\begin{figure*}[t]
  \includegraphics[width=\textwidth]{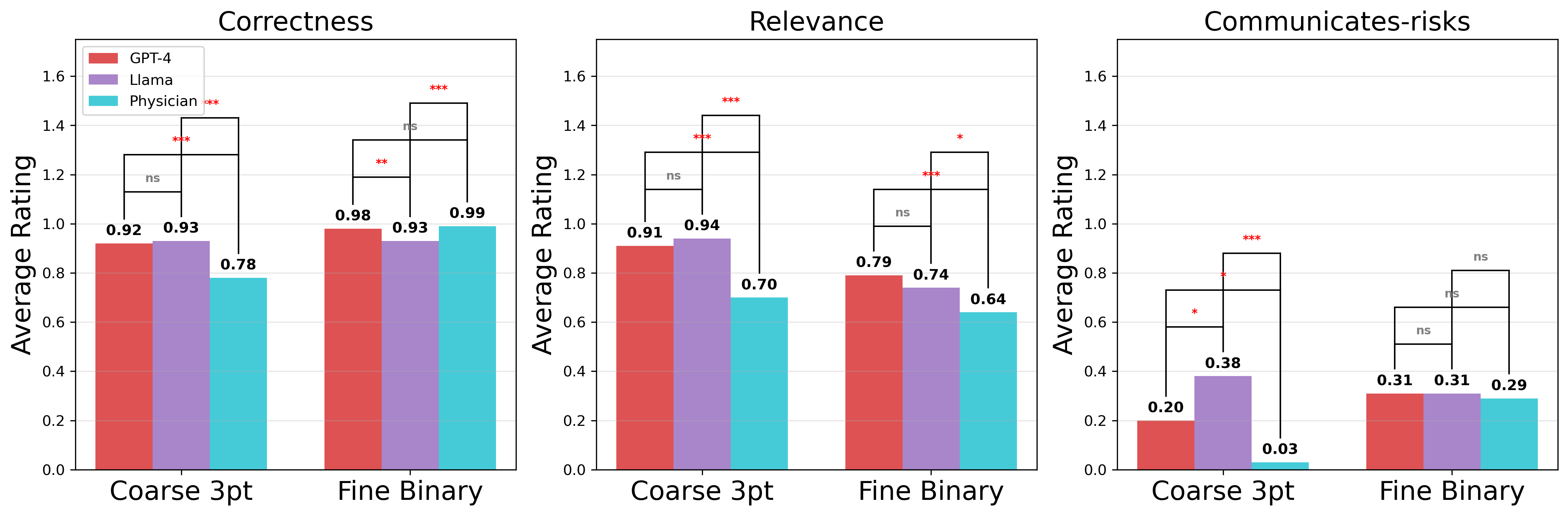}
  \caption{Comparison of system-level average coarse versus fine-grained ratings for correctness, relevance, and communicates-risks. Scores range from 0 to 1, with 1 indicating optimal performance.}
  \label{fig:system}
\end{figure*}

Fine-grained annotations result in comparable system-level ratings (similar range of ratings) for the three annotation dimensions, but produces different system rankings (Figure \ref{fig:system}). GPT-4's and Llama-3.1-Instruct-405B's answers receive near-perfect ratings for correctness in both the coarse (0.92 and 0.93) and fine-grained (0.98 and 0.93) settings. However, where this minimal difference in performance between the two system is not statistically meaningful in the coarse setting, it is significant in the fine-grained one, where GPT-4 shows better performance. Furthermore, physician answers receive much lower correctness ratings in the coarse setting (0.78) than in the fine-grained one (0.99).

Physician answers are consistently shorter and more to the point than model-generated answers, offering limited elaboration (Table~\ref{tab:token_stats}). We hypothesize that this biases annotators toward rating model-generated answers higher, despite comparable correctness, as length may be conflated with informational content, and the style of information presentation used by models may be preferred by users. 
We speculate that by constraining annotators to evaluate one sentence at a time, fine-grained annotations help annotators separate characteristics related to correctness (e.g., accuracy) and relevance (e.g., elaboration), leading to fairer ratings.

For relevance, coarse and fine-grained designs produce different system-level ratings, but the relative system ranking is consistent between the two. All systems receive higher relevance scores in the coarse annotation setting compared to the fine-grained one (0.91, 0.94, 0.70 vs. 0.79, 0.74, 0.64). However, in both settings, GPT-4 and Llama-3.1-Instruct-405B answers are rated statistically similarly, while physician answers are rated significantly lower.

For communicates-risks, Llama’s ratings are similar across coarse and fine-grained settings (0.38 vs. 0.31), whereas GPT-4 and physician outputs show substantially lower communicates-risks ratings for coarse annotations (0.20 and 0.03) than for fine-grained annotations (0.31 and 0.29). Similar to correctness, the lower ratings for physician answers disappears in the fine-grained annotation setting. Overall, however, all models and physicians are rated poorly in terms of communicates-risks across both settings.

Our ratings are consistent with prior work showing that state-of-the-art LLMs can perform at or above physician level on measures of correctness and relevance, while struggling with communicates-risks. \cite{ayers_comparing_2023, kanithi_medic_2024, hosseini_benchmark_2024}.

\textbf{\textit{Recommendation}}: Our findings suggest that fine-grained annotations should generally be preferred for producing system ratings and rankings, particularly for correctness, because they may reduce biases introduced by answer length and presentation style and yield fairer comparisons across systems. Coarse-grained annotations are adequate for some dimensions such as relevance. More work is needed to assess whether model-generated answers provide satisfactory risk information.

\paragraph{Prompting with fine-grained annotation instructions does not consistently improve agreement between LLM-as-judge and expert annotators.}

\begin{figure}[t]
  \includegraphics[width=\columnwidth]{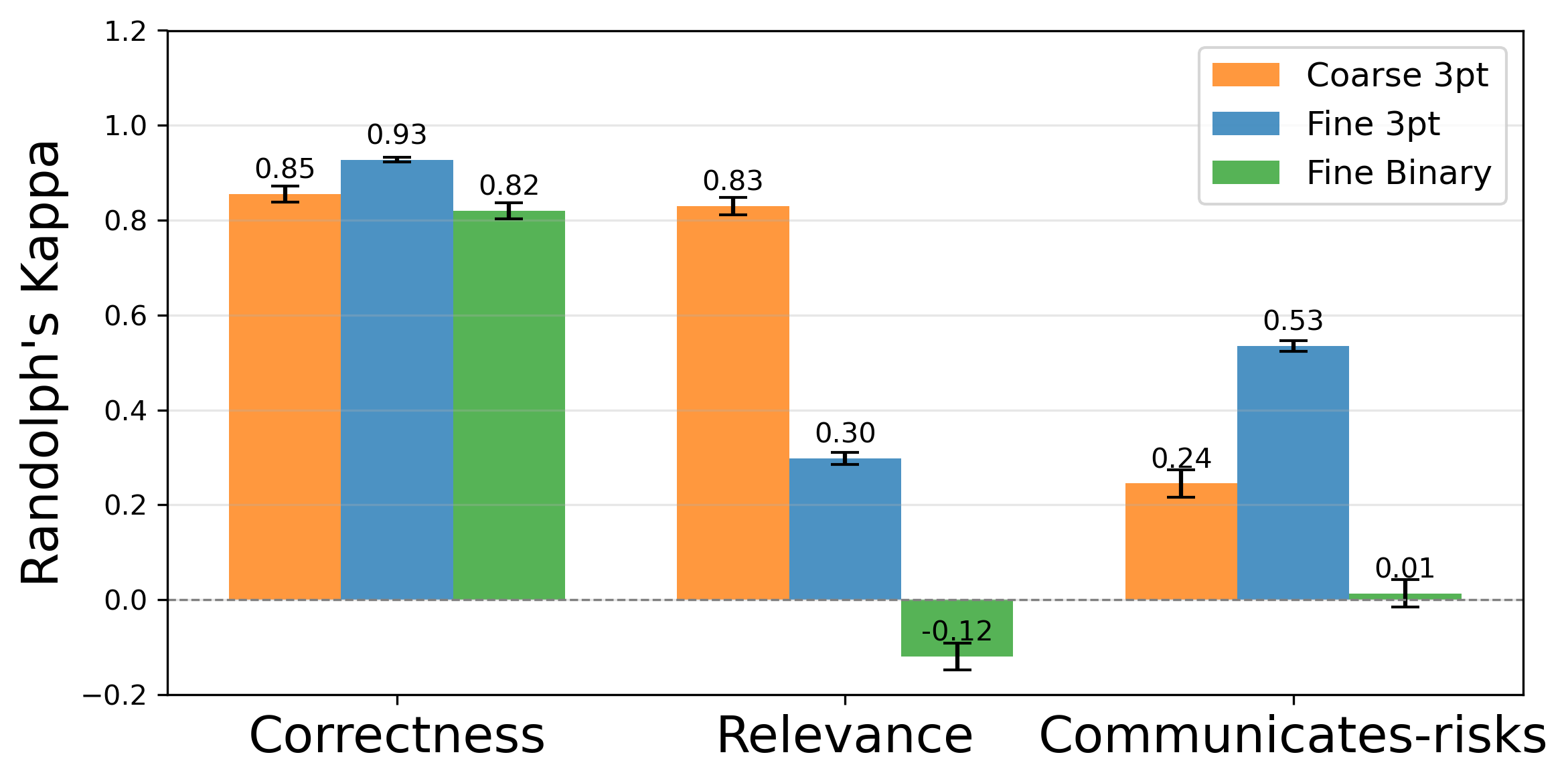}
  \caption{Average inter-annotator agreement (Randolph's $\kappa$) between average expert ratings and LLM-judge ratings for correctness, relevance, and communicates-risks. 
  }
  \label{fig:llm-as-judge}
\end{figure}


Figure~\ref{fig:llm-as-judge} shows IAA between GPT-4o (our judge model) and expert annotators in the coarse and fine-grained evaluation settings. We observe that prompting with fine-grained instructions does not consistently improve human-LLM agreement across rating schemes. Fine-grained 3-point ratings enhance IAA for correctness and communicates-risks, though the IAA remains lower across both settings, while they decrease it for relevance. In contrast, fine-grained binary ratings reduce IAA across all dimensions. Even so, when prompted with our annotation instructions, LLM-as-judge can achieve agreement with experts that is comparable to—or even higher than—expert-expert agreement for correctness and relevance.


\textbf{\textit{Recommendations}}: We do not propose LLM-as-judge as a replacement for expert evaluation, but as a supplement when expert resources are constrained. If leveraging LLM-as-judge to evaluate multi-paragraph clinical QA, fine-grained annotation instructions can improve agreement with expert raters for factual dimensions such as correctness on a 3-point scale, while coarse annotation instructions remain more suitable for dimensions such as relevance. In our experiments, we also find that prompting GPT-4o with our annotations instructions yield LLM-expert IAA that is comparable to, and in some cases exceeds, expert-expert IAA for correctness and relevance. These findings suggest that LLM-as-judge could be effectively used to supplement expert judgments when evaluating these dimensions in annotation studies of clinical QA with constrained resources.

\section{Discussion \& Conclusion}

In this work, we introduce \framework, a framework for evaluating multi-paragraph clinical question answering using both coarse (answer-level) and fine-grained (sentence-level) annotation settings. Through a randomized human annotation study over 300 QA pairs of real patient questions, we find that annotation design matters by dimension: fine-grained annotations improve IAA for factual dimensions like correctness, while coarse annotations remain more reliable for context-dependent dimensions such as relevance. However, our framework did not improve IAA for communicates-risks, which remains low in both fine and coarse annotation settings. This result may be diagnostic of the intrinsic difficulty of this dimension: identifying missing or insufficient risk disclosures requires annotators to reason about what should have been stated in the answer, rather than what is contains. The low agreement suggests that annotation design alone is insufficient for achieving reliable judgments for this type of dimension; future work should explore alternative strategies (e.g., counterfactual prompts, risk checklists).  

We further demonstrate that partial fine-grained evaluation (annotating only three sentences per answer) preserves agreement compared to full fine-grained annotations while reducing annotation effort and cost, making it a practical alterative for resource-constrained studies, particularly when evaluating correctness.

Fine-grained annotations also help mitigate answer length-related biases in system-level evaluations, ensuring that shorter, physician answers are not systematically penalized compared to longer, more elaborated, model-generated responses. Our analysis found that both of the LLMs we tested perform comparably to physicians on the correctness and relevance of answers, though communicates-risks remains a persistent weakness for all models and physicians. This highlights the need for targeted evaluation and improvement in this critical aspect of consumer health QA.

Overall, our findings provide actionable guidance for designing multi-paragraph clinical QA evaluation studies. We recommend employing partial fine-grained annotations to achieve high IAA for correctness while keeping annotator workloads manageable. For relevance, we recommend retaining coarse-level annotations to preserve context. These insights can help researchers and practitioners balance annotation quality, cost, and reliability, ultimately supporting the safe and effective deployment of LLM-powered question-answering systems in patient care.

\section*{Limitations}
\label{sec:limit}
Our dataset is relatively small and consists of only questions on primary care topics, which may limit the applicability of our findings to clinical questions about conditions that require specialized care or to diverse patient populations who might experience such conditions. While the dataset we use in this study is smaller than other datasets with free-text response answers, we chose this dataset because it contains real patient questions and reference answers curated by a team of physicians, providing a high-quality testbed for studying evaluation methods in a realistic consumer health QA scenario. It is also easily accessible, allowing for the reproducibility of our study. To this set we add LLM-generated answers, a codebook for evaluation, expert ratings collected in two settings, IAA measurements, and LLM-as-judge ratings, providing a helpful expansion to the existing literature. We provide a comprehensive comparison to prior work in Table \ref{tab:dataset_comparison}.

Second, expert annotators were recruited on UpWork. While we did request medical credentials and verify them to the best of our ability, there is still a possibility that the credentials our annotators provided are outdated, overstated, or falsified.

Third, we recruited only six expert annotators, which constrains the variability in perspectives and may affect the robustness of inter-annotator agreement (IAA) and intra-rater reliability (IRR) estimates. This choice reflects the practical constraints of working with scarce and costly medical expertise. Our annotation study employs more annotators than other comparable evaluation studies in the medical QA domain \cite{ben_abacha_question-entailment_2019, jin_better_2023, ayers_comparing_2023, singhal_large_2023, khasentino_personal_2025, yim_morqa_2025}, and the results we report achieve statistical significance. Our study design also mirrors the resource-limited conditions under which clinical QA systems are often evaluated in practice, and our recommendations are intended to be useful in such settings.

Fourth, sentence-level annotations may not guarantee coverage of incorrectness between sentences (e.g. dosages), although our findings show that sentence-level annotations produce safety ratings equivalent to passage-level ratings. 

Furthermore, the sample we collect to measure IRR is small, potentially limiting the precision of reliability assessments across dimensions. Among this small sample, we find that fine-grained annotations do not achieve higher IRR across all dimensions. Details and interpretation provided in Appendix~\ref{sec:IRR}. This result highlights the need for larger follow-up studies to confirm whether the trends we observe in IAA extend to individual annotator self-consistency.

Finally, our evaluation focuses on only two widely used language models, GPT-4 and Llama-3.1-Instruct-405B. These models were chosen because they have been documented as performing well on clinical QA, and we limit the number of models in our study in order to obtain more annotations over a larger set of questions. Nonetheless, our findings may not extend to other models or to future versions of these models with different capabilities. Future work should investigate larger datasets, more annotators, and a broader range of models to validate and extend these findings.

\section*{Ethical Considerations}
The design of our human annotation study was submitted to our university's Institutional Review Board (IRB), which determined that no IRB approval or exemption was required, as the study does not meet the regulatory definition of human subjects research. All annotators recruited through Upwork were fairly compensated at a rate of 25 USD per hour, consistent with their suggested hourly wage. We provided participants with sufficient time to complete their annotations, offered flexible deadlines to accommodate personal or health-related needs, and remained available to address feedback and questions throughout the study.


\bibliography{longaQAeval, custom}

\appendix

\section{Example Questions from K-QA}
\label{sec:kqa-examples}

Table~\ref{tab:qa_examples} presents representative consumer health questions from the K-QA dataset alongside corresponding answers provided by the physician (from the dataset), GPT-4, and Llama-3.1-Instruct-405B. These examples illustrate a diversity of question types and differences in style, length, and content among human- and model-generated responses. \textbf{These examples are provided solely for research and discussion purposes and should not be interpreted as medical advice.}

\begin{table*}[t]
\centering
\small
\renewcommand{\arraystretch}{1.2}
\setlength{\tabcolsep}{6pt}
\begin{tabular}{lL{136mm}}
\toprule
\textbf{Question} & \textit{Is accutane like an antibiotic and would it affect a baby if I got pregnant?} \\
\midrule
\textbf{Physician} & Isotretinoin (Accutane) is a form of vitamin A that helps to reduce the amount of oil produced by your skin's oil glands, but it's not an antibiotic. This drug is used to treat severe acne that has not responded to other treatments. It's important to note that this medication can cause severe birth defects or even fetus death if taken during pregnancy. It's crucial to have a negative pregnancy test before starting Isotretinoin and to use two forms of birth control throughout the treatment. If you're considering becoming pregnant or are already pregnant, you should speak to your healthcare provider about alternatives. \\
\midrule
\textbf{GPT-4} & Accutane, also known as isotretinoin, is not an antibiotic. It is a form of vitamin A that helps the skin renew itself more rapidly and is used primarily to treat severe nodular acne. Unlike antibiotics, which eliminate bacteria causing infections, isotretinoin works by reducing the amount of oil released by oil glands in the skin and helping the skin renew itself. Isotretinoin can have serious side effects and is known to cause severe birth defects if taken during pregnancy. It is critical to avoid pregnancy while taking this medication. Women of childbearing age are required to use two forms of birth control and undergo regular pregnancy tests before, during, and after the treatment with isotretinoin. It is vital to consult with a healthcare provider to discuss the risks and necessary precautions while on this medication. \\
\midrule
\textbf{Llama-3.1} & Accutane, also known as isotretinoin, is not an antibiotic but a powerful oral retinoid medication used to treat severe acne, including nodular and conglobate acne, which are resistant to other treatments. Accutane can cause severe birth defects if taken during pregnancy. It is contraindicated in pregnant women, and the FDA mandates the iPLEDGE program to monitor and prevent use during pregnancy. This program requires patients to commit to using two forms of birth control and undergo regular pregnancy tests to prevent pregnancy while taking Accutane.If you are taking Accutane and become pregnant, you should immediately stop taking the medication and inform your doctor. Due to the high risk of birth defects, your doctor may discuss options for ending the pregnancy. It is essential to carefully consider the risks and benefits of taking Accutane, especially if you are a woman of childbearing age. \\
\midrule
\textbf{Question} & \textit{what causes ringworm ?} \\
\midrule
\textbf{Physician} & Ringworm is a contagious skin infection caused by fungi. It is characterized by a circular, red, and itchy rash and can affect the skin, hair, and nails. The infection can be transmitted through contact with people or pets, as well as through shared spaces like locker rooms and shower stalls.Treatment usually consists of topical antifungal drugs. \\
\midrule
\textbf{GPT-4} & Ringworm, also known as tinea, is a fungal infection of the skin caused primarily by dermatophytes such as Trichophyton, Microsporum, and Epidermophyton. These fungi thrive in warm, moist environments, making the skin a favorable site for infection. It spreads through direct contact with infected individuals, animals, or contaminated objects like towels, clothing, and surfaces. The infection is characterized by a circular, red, itchy rash that may be scaly or have a raised border, resembling a ring. Proper hygiene and avoidance of sharing personal items can help prevent ringworm. Treatment typically involves antifungal medications applied topically or taken orally depending on the severity and location of the infection. \\
\midrule
\textbf{Llama-3.1} & Ringworm, also known as tinea, is a fungal infection that affects the skin, hair, and nails. It is caused by a group of fungi called dermatophytes, which include Trichophyton, Microsporum, and Epidermophyton. These fungi thrive in warm, moist environments and can infect humans through direct contact with an infected person, animal, or contaminated object. Ringworm can also be spread through indirect contact, such as walking barefoot in areas where the fungus is present, or by sharing personal care items like towels or clothing. In some cases, ringworm can be contracted from infected animals, such as cats or dogs. The most common causes of ringworm include:
\begin{itemize}[noitemsep, topsep=2pt, leftmargin=14pt]
    \item Direct contact with an infected person or animal
    \item Indirect contact with contaminated objects or surfaces
    \item Sharing personal care items
    \item Walking barefoot in areas where the fungus is present
    \item Poor hygiene and skin care
\end{itemize}
Symptoms of ringworm typically appear within 4-10 days of exposure and can include a circular, itchy rash, redness, and scaling. Treatment for ringworm usually involves antifungal medications, either topical or oral, and good hygiene practices to prevent further spread of the infection. \\
\bottomrule
\end{tabular}
\caption{Example patient questions and answers from physicians, GPT-4, and Llama-3.1-Instruct-405B. Text is lightly edited for formatting. These examples are provided for research and should not be taken as medical advice.}
\label{tab:qa_examples}
\end{table*}

\section{Annotation Dimensions}
\label{sec:dimensions}

Extended definitions for our evaluation dimensions are as follows:
\begin{enumerate}[itemsep=0pt, topsep=2pt, leftmargin=14pt]
    \item The answer aligns with current medical knowledge. This dimension considers whether the answer:
    \begin{itemize}[noitemsep, topsep=0pt, leftmargin=10pt]
        \item Provides information based on existing scientific evidence
        \item Expresses certainty in line with the current state of medical knowledge
        \item Does not contain contradictions or logical flaws
    \end{itemize}

    \item The answer addresses the specific medical question. This dimension considers whether the answer:
    \begin{itemize}[noitemsep, topsep=0pt, leftmargin=10pt]
        \item Directly addresses all the concerns included in the question
        \item Provides requested details and context
        \item Does not contain information unrelated to the question
    \end{itemize}

    \item The answer communicates contraindications or risks. This dimension considers whether the answer:
    \begin{itemize}[noitemsep, topsep=0pt, leftmargin=10pt]
        \item Explains risks in clear, accessible language
        \item Details all potential consequences
    \end{itemize}
\end{enumerate}

\section{Feedback Survey (modeled after NASA-TLX)}
\label{sec:nasatlx}

Annotators complete a feedback survey at the end of our study. The following questions are adapted from the NASA-TLX and asked for each of the coarse- versus fine-grained annotation settings. Each question is answered on a 7-point scale:

\begin{itemize}[noitemsep, topsep=0pt, leftmargin=12pt]
    \item How mentally demanding was the task?
    \item How physically demanding was the task?
    \item How hurried or rushed was the pace of the task?
    \item How successful were you in accomplishing what you were asked to do?
    \item How hard did you have to work to accomplish your level of performance?
    \item How insecure, discouraged, irritated, stressed, and annoyed were you?
\end{itemize}



\section{Feedback Survey Responses}
\label{sec:feedback-survey}

These are some quotes from annotators' open-text responses to the question ``Which annotation design do you prefer?''

\begin{quote}
     \textbf{Quote from Annotator A:} \textit{Presenting the response in a well-structured paragraph ensures that all relevant information is covered comprehensively.}
\end{quote}

\begin{quote}
    \textbf{Quote from Annotator B:} \textit{Sentence level or ''highlighted text'' was kind of tricky because you need to put the highlighted text in context as well.}
\end{quote}

\begin{quote}
    \textbf{Quote from Annotator C:} \textit{I will prefer sentence-level annotation. Though it was more mentally demanding for me but there were more depth to the annotation and i felt i was able to annotate questions better [...]}
\end{quote}

\begin{quote}
    \textbf{Quote from Annotator D:} \textit{it was easier to interpret levels [Likert-scale values] in sentence type especially the third question [communicates-risks].}
\end{quote}

\section{Intra-rater reliability (IRR)}
\label{sec:IRR}

\begin{figure}[t]
  \includegraphics[width=\columnwidth]{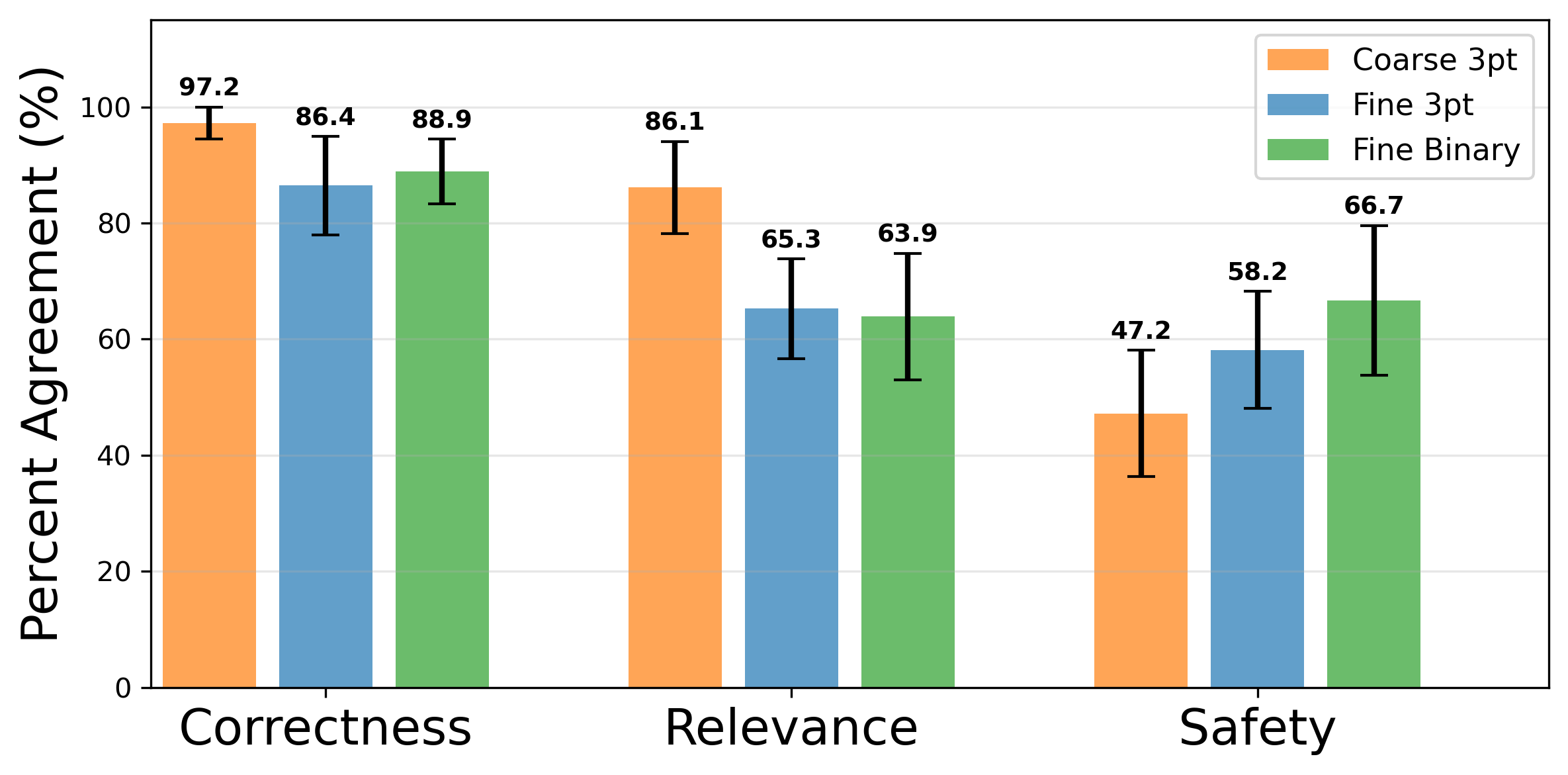}
  \caption{Average intra-rater reliability (IRR) (percent agreement) across annotators for correctness, relevance, and communicates-risks.}
  \label{fig:irr}
\end{figure}

Fine-grained annotations achieve higher IRR for only one of three dimensions (Figure \ref{fig:irr}). IRR was measured for each annotator using percentage agreement and was then averaged for each annotator. For correctness and relevance, reliability is lower in the fine-grained settings (86.4 and 63.9, respectively) compared to the coarse 3-point settings (97.2 and 86.1). Communicates-risks shows an increase in reliability in the fine-grained setting (66.7) compared to the coarse setting (47.2).

These results suggest that fine-grained annotation can improve consistency for highly subjective dimensions such as communicates-risks. In contrast, it does not appear to increase reliability for more objective, factual dimensions like correctness, although reliability in this setting remains satisfactory. For relevance, reliability decreases under the fine-grained setting, mirroring the drop observed in IAA. Our findings are based on a limited sample of question–answer pairs, and further work is needed to validate these observations.


\end{document}

%% file: commands.tex
\definecolor{lightblue}{RGB}{212, 235, 255}
\definecolor{orange}{RGB}{255, 105, 0}
\definecolor{lightgreen}{RGB}{177, 231, 171}
\definecolor{lightyellow}{RGB}{255, 255, 148}
\definecolor{darkgreen}{RGB}{50, 200, 50}
\definecolor{darkyellow}{RGB}{230, 200, 0}

\newcolumntype{P}[1]{>{\centering\arraybackslash}p{#1}}
\newcolumntype{L}[1]{>{\raggedright\let\newline\\\arraybackslash\hspace{0pt}}m{#1}}
\newcolumntype{R}[1]{>{\raggedleft\arraybackslash}p{#1}}

%% file: 02-related-work.tex
\section{Related Work}

\paragraph{Annotation Rubrics}
Human evaluation is the gold standard for assessing clinical QA, as it captures qualities such as factual correctness, safety, and contextual appropriateness that automated metrics often miss~\citep{singhal2023large,singhal2025toward,yim_morqa_2025, khasentino_personal_2025}. Recent benchmarks incorporate human evaluation but rely on broad categorical rubrics without detailed annotation guidance: MultiMedQA~\citep{singhal2023large} and MedQA~\citep{jin2021disease} use three-level scales for correctness and harm, while HealthBench~\citep{arora2025healthbench} and MEDIC~\citep{kanithi2024medic} apply general Likert-style ratings at the answer level. Such underspecified schemes reduce clarity, inter-annotator agreement, and reproducibility.
Moreover, annotator agreement is seldom reported: among 14 expert-annotated studies, only four report IAA \cite{singhal_towards_2023, ben_abacha_question-entailment_2019, johnson_assessing_2023, hosseini_benchmark_2024}, and just one \cite{ayers_comparing_2023} releases annotations. This lack of transparency hinders the development of automatic metrics and systematic comparison between human and LLM judgments.
To address these shortcomings, we introduce \framework, a structured framework with explicit definitions, sub-criteria, and examples, enabling more consistent and actionable evaluation of clinical QA. In addition, we release the human and LLM-generated annotations from our study, along with the corresponding IAA.


\paragraph{Annotation Granularity}
Most prior work in clinical QA uses answer-level evaluation, where annotators assign a single score to the entire response~\citep{singhal2023large,arora2025healthbench,kanithi2024medic,guo2024appls,yim_morqa_2025, mallinar_scalable_2025}. While simple, this approach obscures mixed-quality content. Long answers may include both correct and incorrect statements, safe and unsafe advice, or relevant and irrelevant details. Collapsing these into one score makes error localization difficult, increases subjectivity, and reduces inter-annotator agreement~\citep{castilho2020same,popovic2021agree,krishna_longeval_2023}. In contrast, sentence-level evaluation assesses each unit independently, offering clearer error signals, higher reliability, and more actionable feedback for system development. 